\RequirePackage{fix-cm}
\documentclass[smallcondensed]{svjour3}       % onecolumn (second format)
\smartqed  % flush right qed marks, e.g. at end of proof
\usepackage{graphicx}

\usepackage{rotating}
\usepackage{fancybox}
\usepackage{enumerate}
\usepackage{algorithm}
\usepackage{algpseudocode}
\usepackage{color}
\usepackage[framemethod=tikZ]{mdframed}
\usepackage{comment}
\usepackage{amssymb}
\usepackage{amscd,bm,amsbsy}
\usepackage{setspace}
\usepackage{amsmath}
\usepackage{amsfonts}
\usepackage[round]{natbib}
\newcommand{\by}{\boldsymbol{y}}

\newcommand{\bD}{\boldsymbol{D}}

\newcommand{\bW}{\boldsymbol{W}}

\newcommand{\bx}{\boldsymbol{x}}

\newcommand{\bomega}{\boldsymbol{\omega}}

\newcommand{\TP}{\mathrm{TP}}
\newcommand{\TN}{\mathrm{TN}}
\newcommand{\FP}{\mathrm{FP}}
\newcommand{\FN}{\mathrm{FN}}
\newcommand{\Prec}{\mathrm{Prec}}
\newcommand{\Rec}{\mathrm{Rec}}
\newcommand{\Fmax}{F_{\max}}
\usepackage{soul}

\newcommand{\inda}{\hspace{3mm}}

\newcommand{\indc}{\hspace{12mm}}
\newcommand{\indd}{\hspace{16mm}}

\begin{document}

\title{
%Combining Negative Selection with \Rev{Imbalance-aware} Classification  for Protein Function Prediction\\
%\Rev{\indb POSSIBILE nuovo TITOLO:\\
Positive and Unlabeled Learning through Negative Selection and Imbalance-aware Classification
%\textsl{OPPURE}\\
%Partially Supervised learning through Negative Selection and \Rev{Imbalance-aware} Classification\\
%\textsl{OPPURE}\\
}

\subtitle{}


\author{Marco Frasca         \and
        Nicol\`o Cesa-Bianchi %etc.
}

\authorrunning{M. Frasca and N. Cesa-Bianchi} % if too long for running head

\institute{Marco Frasca \at
              Dipartimento di Informatica, Universit\`a degli Studi di Milano, Milan, 20135, Italy\\
              Tel.: +39-02-50316295\\
%              Fax: +123-45-678910\\
              \email{frasca@di.unimi.it}           %  \\
           \and
           Nicol\`o Cesa-Bianchi \at
              Dipartimento di Informatica, Universit\`a degli Studi di Milano, Milan, 20135, Italy\\
              Tel.: +39-02-50316280\\
%              Fax: +123-45-678910\\
              \email{cesa-bianchi@di.unimi.it}
}

\date{Received: 24/01/2019 / Accepted: date}

\maketitle

\begin{abstract}
Motivated by applications in protein function prediction, we consider a challenging supervised classification setting in which positive labels are scarce and there are no explicit negative labels. The learning algorithm must thus select which unlabeled examples to use as negative training points, possibly ending up with an unbalanced learning problem. We address these issues by proposing an algorithm that combines active learning (for selecting negative examples) with imbalance-aware learning (for mitigating the label imbalance). In our experiments we observe that these two techniques operate synergistically, outperforming state-of-the-art methods on standard protein function prediction benchmarks.

\end{abstract}

\section{Introduction}
\label{intro}
Learning from positive and unlabeled data is a classification setting in which classes have no explicit negative labels ---see, e.g., \citep{elkan2008learning}. An important real-world instance of this problem is the automated functional prediction (AFP) of proteins~\citep{zhao2008gene,Radivojac13}. Indeed, public repositories for protein functions ---e.g., the Gene Ontology~\cite{GO00}--- rarely store ``negative'' annotations of proteins to functions. Another example is the disease-gene prioritization~\citep{Moreau12}, as only genes involved in the disease etiology are usually recorded ---see for instance the Online Mendelian Inheritance in Man (OMIM) database~\citep{Amberger11}. In these domains the lack of a positive annotation for a given class does not necessarily imply that the data point is a true negative for that class. On the contrary, further investigations may result in new annotations being subsequently added to previously unannotated data points.

The absence of explicit negative labels is typically handled by means of a strategy for selecting unlabeled examples that are then used as negative training examples for the class at hand. Since ---as noted earlier--- not all unlabeled instances are true negatives, the task of learning from positive and unlabeled data is generally harder than standard supervised learning. The scarcity of positive examples, which is widespread in most datasets for this setting, just makes things worse, as the only way to increase the size of the training set is by adding negative examples, which eventually leads to an imbalance between positives and negatives.

In this work we propose a novel approach to learning from positive and unlabeled data based on combining active learning (for the negative selection) with imbalance-aware classification (for mitigating the label imbalance). 
Unlike traditional active learning, which is typically used to select training examples from a set of unlabeled data, our active learning approach focuses on the selection of negative examples from a set of unlabeled data that are mostly negative for the class under consideration. The intuition is that the ability of active learning to focus on the most informative examples can be exploited to filter out unlabeled data which are not good negative training points, or possibly not even true negative points. In practice, however, the benefit of selecting negative examples through active learning can be neutralized by the degradation of performance caused by the training imbalance. This can be fixed through the use of an appropriate imbalance-aware learning technique. In our experiments we observe that ---when combined--- active learning and imbalance-aware training operate synergistically, delivering a significant increase of performance on standard AFP benchmarks. Overall, the proposed framework is composed of two phases:
\begin{enumerate}
\item negative training examples are selected through active learning;
\item a supervised classifier, handling the data imbalance, is then learned on the training set containing the available positives and the negatives selected in the previous phase.
\end{enumerate}
We deal with the rarity of positive examples using imbalance-aware learning algorithms, focusing in particular on learners that can be trained both in active and passive learning settings. In the first phase, we incrementally refine an initial model trained on a small seed by running the learner in active mode. This process goes on until a fixed budget of negative instances is reached. In the second phase, a new model is trained by running the learner in passive mode on the training set built in the first phase. A relevant feature of our approach is that, unlike existing approaches~\citep{Mostafavi09,NOGO14,Youngs13}, our method can be applied even in settings where no hierarchy of labels is available.

With regard to the AFP application, our method outperforms state-of-the-art baselines in predicting thousands of GO functions for the proteins of \textit{S.~cerevisiae} and \textit{Homo sapiens} organisms in a genome-wide fashion. 

\section{AFP Related Works}
Computational methods play a central role in annotating the functions of large amounts of proteins delivered by high-throughput technologies. Despite the encouraging results achieved by these methods,
many functions still have a very low number of verified protein annotations, leading to a pronounced imbalance between annotated and unannotated proteins.
Several solutions have been proposed in the last decades for the AFP problem, including those presented in the recent international challenges CAFA1~\citep{Radivojac13} and CAFA2~\citep{CAFA2}, which collected dozens of algorithms proposed by numerous research groups evaluating them on a common set of target proteins. Such methods can be roughly categorized into three groups: (1) graph-based approaches~\citep{Marcotte99,Oliver00,Schwikowski00,Vazquez2003,Sharan07,Mostafavi08,Frasca15homcat}, attempting to transfer functional evidence in a semi-supervised manner from annotated nodes in the protein graph; (2) multitask and structured output algorithms~\citep{Mostafavi09,Sokolov10,Sokolov13,FrascaMTLP,Feng17}, predicting multiple GO functions at the same time by taking into account their hierarchical relationships; (3) hierarchical ensemble methods~\citep{Obozinski08,Guan08,Lan13,Vale14c,Robinson15}, recombining post-prediction inferences in order to respect the hierarchical constraints.

Despite the diversity of these approaches,
to the best of our knowledge no approach has been proposed yet to simultaneously handle the problems of data-imbalance and negative selection. 
Imbalance-aware techniques for AFP, including the cost-sensitive learning, were studied in a few papers, with generally promising results~\citep{Bertoni11,CB09,GarciaLopez13}. A few more works in the same context investigated the problem of negative selection~\citep{Mostafavi09,Youngs13,NOGO14,Frasca17iwbbio}, mostly exploiting the GO structure to select negatives. 
\section{Methods}
\label{sec:alg}

\paragraph{Preliminaries.}
%\label{sub:prel}
Following the standard encoding of biological data, we use a matrix-based data representation in which instances are represented by a connection matrix $\bW$, where the row $\bW_{i,\cdot}$ is viewed as a feature vector for instance $i$. More specifically, our data are represented as follows:
\begin{enumerate}%[topsep=0pt,parsep=0pt,itemsep=0pt]
\item a network of instances, represented as an undirected weighted graph $G = \langle V, \bW\rangle$, where $V \equiv \{1,\ldots,n\}$ is the set of nodes and $\bW$ is a $n\times n$ matrix whose entries $W_{ij}\in[0,1]$ encode some notion of similarity between nodes (with $W_{ij}=0$ when nodes $i,j$ are not connected);
\item a labeling for a given class, described by the binary vector
\[
	\by=(y_1,y_2, \ldots, y_n)\in \{0,1\}^n
\]
where $y_{i} = 1$ if and only if node $i$ is positive for that class.
\end{enumerate}
Let $V_+ \equiv \{i \in V \mid y_{i}=1\}$ and $V_- \equiv V\setminus V_+ \{i \in V \mid y_{i}=0\}$ be the subsets of positive and negative nodes, respectively. Note that we allow the labeling to be highly unbalanced, that is $|V_+| \ll |V_-|$. In addition, the labeling is known only for a subset $S\subset V$ of nodes and unknown for the remaining $U \equiv V \setminus S$. The problem is to infer the labeling of nodes in $U$ using the training labels $S$ and the connection matrix $\bW$. 

\paragraph{Negative Selection.} 
As mentioned in the introduction, the problem of learning from positive and unlabeled data is hard because negative examples $V_-$ are not well defined, and a nonpositive example might be either a negative or a positive subsequently annotated in light of future studies. This makes the selection of informative negatives among the nonpositive examples a central issue for learning accurate models.

In our setting, considering nonpositive instances as negative implies that some negative labels are \textsl{noisy}, in the sense that they might turn out to be positive in the future. In order to take this issue into account, and in line with previous works~\citep{NOGO14}, the noisy labels are characterized by considering two different temporal releases of the labels for a given class. We denote these labelings with the $n$-dimensional vectors $\by$ and $\overline\by$, assuming $\by$ is the older one. The two releases form a {\em{temporal holdout}}: the labeling $\overline\by$ can be used to verify the quality of predictions made by models learned using labeling $\by$. This is addressed by the following definitions: for a given class, let $V_{--}= \big\{i\in V \mid y_{i}=0 \wedge \overline y_{i}=0 \big\}$ be the set of negative instances whose label did not change, and $V_{-+}= \big\{i\in V \mid y_{i}=0 \wedge \overline y_{i}=1 \big\}$ be the set of instances with noisy labels. We go back to this issue in Section~\ref{sec:results}, where we investigate the behaviour of our models on noisy instances.

\subsection{Active learning for negative selection}\label{sub:AL}
Let $S_+ = S \cap V_+$ and $S_- = S \cap V_-$ be, respectively, the training sets of positive and negative (nonpositive) proteins for a given class. Given a budget $0 < B < |S_-|$, our goal is to select a subset of $B$ negative examples $\widehat S_-\subset S_-$ in order to maximize the performance of the classifier trained using the examples $S_+ \cup \widehat S_-$. 
%Due to the scarcity of annotated proteins, we include all available positives in the training set $L$, that is $L = S_+ \cup \hat S_-$. 

We address this problem using an active learning (AL) strategy. AL is typically used to save labeling cost by allowing the learner to choose its own training data~\citep{settles2012active}. In pool-based AL, the learner receives a label budget $B$ and then selects $B$ instances to be labeled from a large set of unlabeled data. A standard pool-based AL strategy is to obtain the label of those points which the current model classifies with least confidence ---see, e.g., \citep{tong2001support}. Our setting is slightly different because we want the AL algorithm to select points to which we assign negative labels. Nevertheless, we may exploit AL to pick out the ``most informative'' negative points for training our model. 

The pseudocode of our negative selection procedure in supplied in Algorithm~\ref{algAL}.
%%%%%%%%%%%%%%%%%%%%%%%%%%%%%%
\newcommand{\com}[1]{\Comment{\color{gray}#1\color{black}}}
\begin{algorithm}[t]
\caption{Negative examples selection - procedure template}\label{algAL}
\begin{algorithmic}[0]
	\Statex 
	\Statex \textbf{Input:} $G=\langle V,\bW\rangle$, \inda input graph\\
			\indc $S_+$, \indd set of positive nodes\\ 
            \indc $S_-$, \indd set of non positive nodes\\
            \indc $\tilde S_- \subset S_-$, \inda\ \ \  initial negative seed\\
            \indc $s$,  \indd\ \  size of active learning selection \\
            \indc $B$,  \indd\ budget of negatives to be selected
	\Statex \textbf{Output:} $\widehat S_- \subset S_-$, with $|\widehat S_-| = B$.\\		
    \Procedure	{}{}
%	\Statex
	\State $t \leftarrow 0$
	\State $S_-(t) 	\leftarrow \tilde S_-$ 
	\State $I(t) \leftarrow S_+ \cup S_-(t)$ %\com{Initial seed}
	\State $C \leftarrow \textsc{LearnClassifier}(G, I(t))$
	\While {$|I(t)| < B$}
        \State $s' \leftarrow \min \{s,\ B-|I(t)|\}$		
		\State $T \leftarrow \textsc{LCP}(C,\ G,\ S_-\setminus I(t),\ s')$		
		\State $S_-(t+1)\leftarrow S_-(t) \cup T$
		\State $I(t+1)\leftarrow I(t) \cup T$
		\State $C \leftarrow \textsc{UpdateClassifier}(C,\ G,\ I(t+1))$
		\State $t \leftarrow t+1$	
	\EndWhile
	\State $\widehat S_- \leftarrow S_-(t)$
	\State \Return $\widehat S_-$
\EndProcedure
\end{algorithmic}
\end{algorithm}
%\paragraph{Active learning procedure (template).}
An initial seed training set $I(0) = S_+ \cup S_-(0)$ is defined, where $S_-(0) \subset S_-$ is selected at random and balanced (i.e., $|S_-(0)|= |S_+|$).\footnote{
A balanced seed training set is empirically the best performing choice.
}
$I(0)$ contains all available positives (since we do not have many of them anyway).
The procedure \textsc{LearnClassifier} learns an initial classifier $C$ using the points in $I(0)$.
In each iteration $t=0,1,\dots$, a new training set $I(t+1)$ is built by adding to $I(t)$ the $s$ instances in $S_-\setminus I(t)$ that are predicted with least confidence by the current classifier (procedure \textsc{LCP}). Then, the classifier $C$ is updated (or retrained) using $I(t+1)$ by the procedure \textsc{UpdateClassifier}. These steps are iterated until $|I(t)| = |S_+| + B$ (budget is exhausted).

We validate this approach using two popular learning algorithms which have natural AL variants. In the following $L$ denotes the training set and we define $L_-=\{i\in L \mid y_i = 0\}$ and $L_+=\{i\in L \mid y_i = 1\}$.

\subsubsection{Active imbalance-aware SVM}\label{subsub:activesvm} 
Given the training instances $(\bx_i, y_i) \in \mathbb{R}^n\times \{-1, 1\}$, the Support Vector Machine (SVM)~\citep{Vapnik95} learns the hyperplane $\bomega^*\in \mathbb{R}^n$ unique solution of the following optimization problem:
\begin{equation}\label{eq:SVM} 
\begin{array}{l}
	{\displaystyle \min_{\bomega \in \mathbb{R}^n} \frac{1}{2}\|\bomega\|^2 +\ C\sum_i\xi_i}
\\
	\text{s.t.} \quad y_i{\bomega}^{\top}\bx_i \geq 1-\xi_i  \quad i \in L
\\
	\quad\hspace*{3mm} \xi_i \geq 0  \hspace{1.57cm} i \in L~.
\end{array}
\end{equation}
In our setting $\bx_i = \bW_{i,\cdot}$ (the $i$-th row of $\bW$).
The \textit{margin} of the instance $i$ is $\big|{\bomega^*}^{\top}\bx_i\big|$.
The most uncertain instance is the one with lowest margin, located closest to the decision hyperplane~\citep{tong2001support}.
Thus, when we implement the AL procedure template with SVM, the instances selected by the procedure \textsc{LCP} are those with the smallest margin.

In order to deal with the scarcity of positives, we use the cost-sensitive SVM of~\cite{Morik99}, in which the misclassification cost has been differentiated between the positive and the negative classes. The corresponding objective function is
\begin{equation}\label{eq:SVM_cs}
\begin{array}{l}
	{\displaystyle \min_{\bomega \in \mathbb{R}^n} \frac{1}{2}\|\bomega \|^2 +\ C_+\sum_{i:y_i=1} \xi_i + C_-\sum_{i:y_i=0} \xi_i}
\\
	\text{s.t.} \quad y_i{\bomega}^{\top}\bx_i \geq 1-\xi_i  \quad i \in L\\
	\quad\hspace*{1mm} \quad \xi_i \geq 0  \hspace{1.57cm} i \in L~.
\end{array}
\end{equation}
The sum over slack variables in~(\ref{eq:SVM}) is split into separate sums over positive and negative training instances, with two different misclassification costs $C_+$ and $C_-$. As suggested by the authors, we set $C_-=1$ and $C_+={|L_-|}/{|L_+|}$. 
In our experiments, we denote with SVM AL the cost-sensitive SVM using active learning for negative selection.
%%%%%%%%%%%%%%%%%%%%%%%%%%%%%%%%%%%%%%%%%%%%%%%%%%%

\subsubsection{Active imbalance-aware RF}
The Random Forests algorithm (RF)~\citep{Breiman01} builds an ensemble of classification trees, where each tree is trained on a different bootstrap sample of $N < |L|$ random instances, with splitting functions at the tree nodes chosen from a random subset of $M < n$ attributes. RF then aggregates tree-level classifications uniformly across trees, computing for each instance $i$ the fraction $p_{i1}$ of trees that output a positive classification.

When we implement the AL procedure template with RF, the instances selected by the procedure \textsc{LCP} are those with highest entropy $H_i$, where
\begin{equation}\label{entropy}
H_i = - p_{i1}\log p_{i1} - (1-p_{i1})\log(1-p_{i1})
\end{equation}
and $p_{i1}$ is computed according to the RF model $C$.

Similarly to~\citep{VanHulse07,Khalilia11}, in this study we use a variant of RF designed to cope with the data imbalance. When RF selects the examples for training a given tree, an instance $i\in L$ is usually selected with uniform probability $p_i = \frac{1}{|L|}$. Here, instead, we draw positive and negative examples with different probabilities,
\[
\label{eq:weighting-scheme}
p_i = \left\{\hspace*{-1mm}
\begin{array}{cl}
	\frac{1}{2|L_+|} & \text{if  $y_i=1$}
	\\
	\frac{1}{2|L_-|} & \text{if $y_i=0$.}
\end{array} \right.
\]
In this way, the probabilities of extracting a positive or a negative example are both $\frac{1}{2}$, and the trees are trained on balanced datasets.
In our experiments, we denote with RF AL the balanced RF using active learning for negative selection.
%%%%%%%%%%%%%%%%%%%%%%%%%%%%%%%%%%%%%%%%%%%%%%%%%%%%%%%%%%%%%%%%5

\section{Results and discussion}
\label{sec:results}
In this section we analyze the empirical performance of our algorithm on predicting the protein bio-molecular functions. We start by describing the datasets and the evaluation metrics, and then we move on to assess the effectiveness of the proposed approach through three different experiments: the study of the impact the parameter $s$ of the Algorithm 1 on the final performance; the evaluation of the performance in the temporal holdout setting; the comparison of our algorithm against an extensive collection of state-of-the-art baselines for predicting protein functions.

\subsection{Automated functional prediction of proteins}
\label{sub:AFP}
AFP is a central and challenging problem of the post-genomic era, involving sophisticated computational techniques to accurately predict the annotations of new proteins and proteomes. As discussed in the introduction, AFP is a prominent example of learning from positive and unlabeled data, in which some of the unannotated instances may be actually positives for specific classes.

We evaluated our algorithm on two datasets: \textit{Homo sapiens} (human) and \textit{Saccaromyces cerevisiae} (yeast). Each dataset consists of a protein network and the corresponding GO annotations. 
Both networks were retrieved from the STRING database, version 10.5~\citep{STRING10}, which already merges many sources of information about proteins. These sources include several databases collecting experimental data, such as BIND, DIP, GRID, HPRD, IntAct, MINT; or databases collecting curated data, such as Biocarta, BioCyc, KEGG, and Reactome. The connection matrix $\bW$ is obtained from the STRING connections $\widehat{\bW}$ after the symmetry-preserving normalization
$\bW = {\bD}^{-1/2}  \widehat{\bW} \bD^{-1/2}$,
where $\bD$ is a diagonal matrix with non-null elements $d_{ii} = \sum_j \widehat W_{ij}$.
As suggested by STRING curators, we set the threshold for connection weights to $700$. The two networks contain $6391$ yeast and $19576$ human proteins.

We considered all the three GO branches. Namely Biological Process (BP), Molecular Function (MF), and Cellular Component (CC). The temporal holdout was formed by considering two different annotation releases: the UniProt GOA releases $69$ (9 May 2017) and $52$ (December 2015) for yeast, and GOA releases 168 (May 2017) and 151 (December 2015) for human. In both releases, we retained only experimentally validated annotations. 

\paragraph{Evaluation framework.}
To evaluate the generalization capabilities of our methods, we used a 3-fold cross validation (CV)
and temporal holdout evaluation (i.e., old release annotations are used for training while new release annotations are used for testing).

\paragraph{Performance measures.}
The performance of our classifier is measured in terms of precision (P), recall (R) and $F$-measure (F). 
Following the recent CAFA2 international challenge~\citep{CAFA2}, we adopted two measures suitable to evaluating instance rankings: the Area Under the Precision-Recall curve (AUPR) and the multiple-label $F$-measure ($\Fmax$). AUPR is a ``per task'' measure, more informative on unbalanced settings than the classical area under the ROC curve~\citep{Saito15}. $\Fmax$ provides an ``instance-centric'' evaluation, assessing performance accuracy across all classes/functions associated with a given instance/protein. More precisely, if we indicate as $\TP_j(t)$, $\TN_j(t)$ and $\FP_j(t)$, respectively, the number of true positives, true negatives, and false positives for the instance $j$ at threshold $t$, we can define the ``per-instance'' {\em{multiple-label precision}} $\Prec(t)$ and {\em{recall}} $\Rec(t)$ at a given threshold $t$ as:
\begin{align*}
\Prec(t) &= \frac{1}{n} \sum_{j=1}^{n} \frac{\TP_j(t)}{\TP_j(t) + \FP_j(t)}
\\
\Rec(t) &= \frac{1}{n} \sum_{j=1}^{n} \frac{\TP_j(t)}{\TP_j(t) + \FN_j(t)}
\label{eq:prec-rec}
\end{align*}
where $n$ is the number of instances. $\Prec(t)$ (resp., $\Rec(t)$) is therefore the average multilabel precision (resp., recall) across instances.
According to the CAFA2 experimental setting, $\Fmax$ is defined as
\begin{displaymath}
\Fmax = \max_t \frac{2 \Prec(t) \Rec(t)}{\Prec(t) + \Rec(t)} 
\label{eq:F}
\end{displaymath}
\subsection{Evaluating the impact of parameter $s$}\label{sub:ALsteps}
\begin{figure*}[t]
\begin{center}
\begin{tabular}{cc}
\hspace{0.95cm} (a)  & \hspace{0.9cm} (b)\\[-2pt]
\hspace{-0cm} \includegraphics [width=0.4\textwidth] {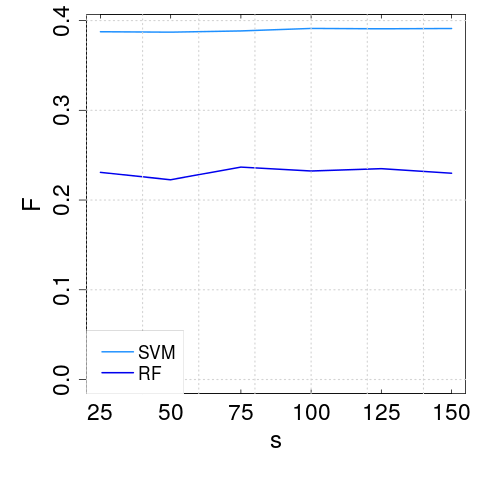}& 
\hspace{-0cm} \includegraphics [angle=-0,width=0.4\textwidth] {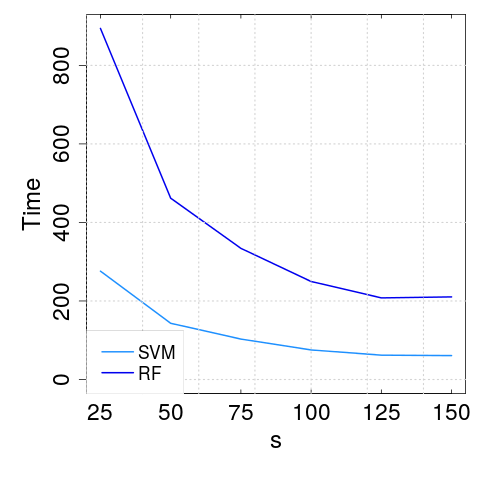}\\
\end{tabular}
\end{center}
\caption{F scores averaged across GO terms (a), and average time in seconds to perform a cycle of $3$-fold CV (b) while varying the active learning parameter $s$.}\label{fig:active.step.eval}
\end{figure*}

We study the trade-off between running time and $F$ values by varying the parameter $s$ (Algorithm~\ref{algAL}) while keeping fixed the budget $B$ of negatives to be selected. Due to the large number of settings under consideration, this experiment only involved yeast data and a subset of GO terms. Specifically, we chose the CC terms with exactly $10$ annotated proteins in the last release (GOA release $69$), for a total of $81$ terms. This choice ensures a minimum of information for learning the model, significantly reduces the number of negatives randomly selected to form the initial seed (which is less than $10$ in the $3$-fold CV), and requires the same number of negatives to be selected through AL (around $B-10$) for each term. We chose the CC branch because, compared to MF and BP branches, it produces the lowest number of terms in the same setting.
We set the budget $B$ to $500$, as a reasonable proportion of the total number of proteins (different values of $B$ showed a similar trend), while $s$ varies in the set $\{25, 50, 75, 100, 125, 150\}$. Values of $s$ lower than $25$ increased the computational burden with a negligible impact of the classification performance, whereas $s>150$ made less significant the contribution of AL (since $B=500$). 
In RF, after tuning on a small subset of labeled data, we decided to use $200$ trees (see Fig.~\ref{fig:active.step.eval}).
%%%%%%%%%%%%%%%%%%%%%%%%%%%%%%%%%%%%%%%%%%%%%%%%%%%%%%%%%
%%%%%%%%%%%%%%%%%%%%%%%%%%%%%%%%%%%%%%%%%%%%%%%

Although in typical AL applications $s$ is set to $1$, these results show that larger values of $s$ do not deteriorate the performance in this setting. Instead, the average F values show almost negligible differences when varying $s$. This behaviour is in agreement with the results in~\citep{Mohamed10}. On the other hand, lower values of $s$ strongly increase the average running time. Due to the above considerations, in the rest of the paper we set $s = 150$, as a good compromise between predictive performance and running time. 

SVMs largely outperform RFs in terms of F in this setting, and are also much faster. This is due to the sparse implementation exploited by the SVMLight library~\cite{Joachims99}, as opposed to the RF implementation provided by the Ranger library~\citep{Wright17}.  Both libraries are written in C language and used within a R wrapper. As a remark, we point out that SVM can be made even faster using other approaches proposed in the literature ---see e.g. \citep{Bordes05,Tsai14}; nevertheless, this is not the main focus in our setting, where data size is relatively small and the quality of solutions does not depend on the specific implementation of the model.

%%%%%%%%%%%%%%%%%%%%%%%%%%%%%%%%%%%%%%%%%%%%%%%%%%%%%%%%%%%%%
\subsection{Assessing the effectiveness of negative selection}\label{sub:afp_cnp}
%%%%%%%%%%%%%%%%%%%%%%%%%%%%%%%%%%%%%%%%%%%%%%%%%%%%%%%%%%%%
\begin{table*}[!t]

\centering
\tiny
\fontsize{6pt}{1pt}
\begin{tabular}{lccccc|cccc||cccc|cccc}
  \hline
{\tt GO} & {\tt B} & {$\rho$} & {\tt P} & {\tt R} & {\tt F} & {$\rho$}& {\tt P} & {\tt R} & {\tt F} & {$\rho$}& {\tt P} & {\tt R} & {\tt F} & {$\rho$} & {\tt P} & {\tt R} & {\tt F}\\ 
  \hline
& &  \multicolumn{8}{c}{\bf Yeast}& \multicolumn{8}{c}{\bf Human}\\ [3pt]  
& &  & \multicolumn{3}{c}{\bf SVM PnoSub}& &\multicolumn{3}{c}{\bf RF PnoSub}& &\multicolumn{3}{c}{\bf SVM PnoSub}& & \multicolumn{3}{c}{\bf RF PnoSub}\\ [1pt]
CC & all  & & 0.457 & 0.539 & 0.480 & & 0.52 & 0.3 & 0.332 &  & 0.174 & 0.168 & 0.158 &  & 0.199 & 0.121 & 0.115\\[1pt]
MF & all & &  0.353 & 0.299& 0.312  & &  0.435 & 0.113 & 0.170  & & 0.33 &0.225 & 0.254 & & 0.298 & 0.089 & 0.125\\[1pt]
BP & all & &  0.48 & 0.447 & 0.445  & &  0.553 & 0.235 & 0.311  & &  0.281 & 0.168 & 0.195 & &0.218 & 0.046 & 0.069\\[1pt]  
\cline{4-6}\cline{8-10}\cline{12-14}\cline{16-18}\\[-3pt]
%%%%%%%%%%%%%%%%%%%%%%%%%%%%%%%%%%%%%%%%%%%%%%%%%%%%%%%%%%%%
%%%%%%%%%%%%%%%%%%%%%%%%%%%%%%%%%%%%%%%%%%%%%%%%%%%%%%%%%%%%%%
%%%%%%%%%%%%%%%%%%%%%%%%%%%%%%%%%%%%%%%%%%%%%%%%%%%%%%%%%%%%%%%
& & & \multicolumn{3}{c}{\bf SVM PrndSub} & & \multicolumn{3}{c}{\bf RF PrndSub}& & \multicolumn{3}{c}{\bf SVM PrndSub} & & \multicolumn{3}{c}{\bf RF PrndSub}\\ [1pt]
CC & 450 &  & 0.197 & 0.790 & 0.310 &  & 0.219 & 0.773 & 0.329 & & 0.087 & 0.387 & 0.133 & & 0.102 & 0.372 & 0.141\\[1pt]
CC & 600  &  & 0.21 & 0.762 & 0.326 &  & 0.254 & 0.726 & 0.363 & & 0.093 & 0.355 & 0.139 & & 0.095 & 0.297 & 0.133\\[1pt]
CC & 750  &  & 0.223 & 0.754 & 0.334 & & 0.289 & 0.685 & 0.386 & & 0.101 & 0.361 &  0.148  & & 0.133 &0.31 &0.162\\[1pt]
\cline{4-6}\cline{8-10}\cline{12-14}\cline{16-18}\\[-3pt]
%%%%%%%%%%%%%%%%%%%%%%%%%%%%%%%%%%%%%%%%%%%
MF & 450 &   & 0.176 & 0.586 & 0.263 &  & 0.133 & 0.531 & 0.216 & & 0.109 &0.469 &0.169  & &0.094 &0.444 &0.148\\[1pt]
MF & 600 &  & 0.195 & 0.547 & 0.282 &  & 0.172 & 0.475 & 0.248 & &0.143 &0.461 &0.203  & &0.115 & 0.402 & 0.172\\[1pt]
MF & 750 &  & 0.207 & 0.521 & 0.297 &  & 0.197 & 0.421 & 0.254 & &0.15 & 0.442 & 0.212 & & 0.151 & 0.378 & 0.204\\[1pt]
\cline{4-6}\cline{8-10}\cline{12-14}\cline{16-18}\\[-3pt]
%%%%%%%%%%%%%%%%%%%%%%%%%%%%%%%%%%%%%%%%%%%%%%%
BP & 450 &   & 0.233 & 0.661 & 0.335 & & 0.201 & 0.636 & 0.295 & &0.079 & 0.385 & 0.121 & & 0.082 & 0.346 & 0.114\\[1pt]
BP & 600 &  & 0.261 & 0.629 & 0.353 &  & 0.259 & 0.602 & 0.338 & &0.088 & 0.357 & 0.13  & &0.111& 0.303& 0.136\\[1pt]
BP & 750 &  & 0.276 & 0.603 & 0.372 & & 0.261 & 0.598 & 0.339 & &0.099 & 0.356 &0.142  & &0.123 & 0.27 & 0.145\\[1pt]
%%%%%%%%%%%%%%%%%%%%%%%%%%%%%%%%%%%%%%%%%%%%%%%%%%%%%%%%%%
& & & \multicolumn{3}{c}{\bf SVM AL} & & \multicolumn{3}{c}{\bf RF AL}& &\multicolumn{3}{c}{\bf SVM AL} & & \multicolumn{3}{c}{\bf RF AL}\\ [1pt]
CC & 450  & 0.601 & 0.449 & 0.619 & 0.495 & 0.606 & 0.505 & 0.453 & 0.413& 0.283 & 0.218 & 0.21 & \underline{\textbf{0.192}} & 0.263 & 0.182 &  0.165 & 0.157\\[1pt]
CC & 600  & 0.765 & 0.488 & 0.578 & 0.519 & 0.67 & 0.638 & 0.438 & 0.424 & 0.352 & 0.202 &0.167 &0.172  & 0.359 & 0.167 & 0.105 & 0.12\\[1pt]
CC & 750  & 0.819 & 0.514 & 0.551 & \underline{\textbf{0.535}} & 0.702 & 0.647 & 0.417 & 0.430 &0.394 &0.191 &0.164 &0.162  & 0.398 & 0.182 & 0.103 & 0.124\\[1pt]
\cline{4-6}\cline{8-10}\cline{12-14}\cline{16-18}\\[-3pt]
%%%%%%%%%%%%%%%%%%%%%%%%%%%%%%%%%%%%%%%%%%%%%%
MF & 450 & 0.412 & 0.343 & 0.331 & 0.317 & 0.388 & 0.312 & 0.294 & 0.227 & 0.433 & 0.304 &0.26 &\textbf{0.255} & 0.406 & 0.338 &0.2 &0.216\\[1pt]
MF & 600 & 0.575 & 0.395 & 0.318 & 0.334 & 0.469 & 0.413 & 0.195 & 0.240 & 0.489 & 0.316 &0.225 & 0.249 & 0.485 & 0.321 & 0.159 & 0.184\\[1pt]
MF & 750 & 0.624 & 0.423 & 0.305 & \underline{\textbf{0.342}} & 0.506 & 0.351 & 0.222 & 0.238 & 0.533 & 0.305 & 0.217 & 0.245 & 0.565 & 0.347 & 0.15 & 0.185\\[1pt]
\cline{4-6}\cline{8-10}\cline{12-14}\cline{16-18}\\[-3pt]
%%%%%%%%%%%%%%%%%%%%%%%%%%%%%%%%%%%%%%%%%%%%5 
BP & 450 & 0.502 & 0.472 & 0.463 & 0.445 & 0.438 & 0.436 & 0.394 & 0.381 &0.226 & 0.222 & 0.199 & 0.193  &0.244 & 0.248 & 0.128 & 0.138\\[1pt]
BP & 600 & 0.588 & 0.513 & 0.449 & 0.457 & 0.559 & 0.476 & 0.337 & 0.373 & 0.329& 0.292 & 0.172 & \underline{\textbf{0.204}}  & 0.331 &  0.277 & 0.089 & 0.122\\[1pt]
BP & 750 & 0.663 & 0.572 & 0.415 & \underline{\textbf{0.468}} & 0.621 & 0.489 & 0.325 & 0.385 & 0.369 & 0.285 & 0.162 & 0.195  & 0.37& 0.254 &0.077 & 0.107\\[1pt]
\hline\hline\\[-5pt]    
\end{tabular}
\caption{Average results of SVM and RF methods for predicting GO terms using the old release of annotations. In bold the best results (underlined when statistically significant).}\label{tab:yeast_svm_rf} 
\normalsize
\end{table*}
The temporal holdout setting is used here both to validate our negative selection procedure and  to verify how our algorithm it behaves on proteins in $V_{-+}$. Models are trained using the older release of annotations $\by$, and their predictions compared with the corresponding labeling $\overline \by$ in the later release.
To this end, the set $V_{-+}$ should contain enough proteins for the 3-fold CV validation, and ---accordingly--- we only considered GO terms with $|V_{-+}|\geq 5$. The final sets of GO terms contain $9$, $24$ and $99$ (CC, MF, and BP, respectively) for yeast, and $139$, $256$ and $707$ for human.  
To equally partition instances $V_{-+}$ among folds, and maintain the stratified setting, each fold is built by uniformly extracting a proportion $\frac{1}{3}$ of instances from $V_+$, $V_{--}$ and $V_{-+}$.  

To better assess the benefit of our active learning selection, we also tested two baselines: ``passive, no subsampling'' (PnoSub), in which we train the passive models (passive variants of the SVM and RF models described in Section~\ref{sec:alg}) on the entire set $S$, and ``passive, random subsampling'' (PrndSub), using $S_+ \cup \bar S_-$ as training set, where $\bar S_-$ is uniformly extracted from $S_-$. 
Furthermore, we also compute the fraction $\rho$ of instances in $V_{-+} \cap S$ selected during active learning.
Table~\ref{tab:yeast_svm_rf} reports the results according to the older release.     
$B=\{450, 600, 750\}$ was used for both organisms, since we experimentally verified that budgets $B>750$ shown negligible differences in the active models performance (data not shown).

In both datasets, compared to the passive variant trained over all available data, random subsampling leads to worse results for SVMs and to better results for RFs, while our strategy always achieves better results.
On the one hand, this confirms that even random subsampling can be useful to reduce the data imbalance and learn more effective models (as it happens for RF); on the other hand, it also shows that random subsampling might discard informative data (like in the case of SVMs), which instead are preserved by our active learning selection. Moreover, unlike PrndSub selection, AL learns more precise classifiers, which is good in settings where positives carry most information. Overall, the improvements obtained when using active learning are statistically significant according to the Wilcoxon signed rank test ($p$-$value < 0.05$)~\cite{Wilcoxon}.

%%%%%%%%%%%%%%%%%%%%%%%%%%%%%%%%%%%%%%%%%%%%%%%%%%%%%%%%%%%%5
Interestingly, proteins $V_{-+}$ tend to be chosen by our AL selection: with a budget representing around the $15\%$ (resp., $5\%$) of negatives in the pool, a fraction $\rho>0.5$ (resp., $\rho>0.3$) of noisy label proteins was selected on yeast (resp., human) data. This reveals that the classifier is typically less certain about the classification of noisy label proteins, which is the desirable behaviour for a negative selection procedure in this setting, since our goal is to select the most informative negatives.
Conversely, if we have to build an oracle answering queries for ``reliable'' negatives, that is negative proteins which likely remain unannotated in the future for that function, our strategy would supply proteins as far as possible from the decision boundary (highest margin) on the negative side. These are precisely the instances whose negative labels the model is most confident about. Consequently, the larger $\rho$, the more effective and coherent our negative selection procedure. As a further analysis, we also tested SVMs and RFs using the same negative selection procedure but without using any imbalance-aware technique: in this case negative selection alone did not significantly help any of the two methods (results not shown), confirming the benefit of the combined action of negative selection and imbalance-aware learning in this context.
%%%%%%%%%%%%%%%%%%%%%%%%%%%%%%%%%%%%%%%%%%%%%%%%%%%%%%%%%%%%%
\begin{table}[t]
\centering
\scriptsize
\begin{tabular}{lcccc|ccc||ccc|ccc}
  \hline
{\tt Branch} & {\tt B} & {\tt P} & {\tt R} & {\tt F} & {\tt P} & {\tt R} & {\tt F} & {\tt P} & {\tt R} & {\tt F} & {\tt P} & {\tt R} & {\tt F}\\ 
  \hline
& &  \multicolumn{6}{c}{\bf Yeast}& \multicolumn{6}{c}{\bf Human}\\ [3pt]
& &  \multicolumn{3}{c}{\bf SVM PnoSub} & \multicolumn{3}{c}{\bf RF PnoSub}&  \multicolumn{3}{c}{\bf SVM PnoSub} & \multicolumn{3}{c}{\bf RF PnoSub}\\ [1pt]
CC & all   & 0.154 & 0.048 & 0.072 & 0.083 & 0.042 & 0.056 & 0.139 &  0.044 & 0.06 &  0.056 & 0.035 & 0.039\\[1pt]
MF & all   & 0.746 & 0.283 & 0.387 & 0.077 & 0.017 & 0.027 & 0.271 & 0.083 & 0.123 & 0.104 & 0.025 & 0.04\\[1pt]
BP & all   & 0.278 & 0.126 & 0.162 & 0.117 & 0.04 & 0.058 & 0.038 & 0.19 & 0.038 & 0.052 & 0.008&  0.014\\[10pt]  
%%%%%%%%%%%%%%%%%%%%%%%%%%%%%%%%%%%%%%%%%%%%%%%%%%%%%%%%%%%%
%%%%%%%%%%%%%%%%%%%%%%%%%%%%%%%%%%%%%%%%%%%%%%%%%%%%%%%%%%%%%%
%%%%%%%%%%%%%%%%%%%%%%%%%%%%%%%%%%%%%%%%%%%%%%%%%%%%%%%%%%%%%%%
& &  \multicolumn{3}{c}{\bf SVM PrndSub} & \multicolumn{3}{c}{\bf RF PrndSub}&  \multicolumn{3}{c}{\bf SVM PrndSub} & \multicolumn{3}{c}{\bf RF PrndSub}\\ [1pt]
CC & 450 	& 0.917 & 0.625 & 0.686 & 0.803 & 0.387 & 0.457 & 0.556 & 0.236 & 0.306 & 0.583 & 0.221 & 0.296\\[1pt]
CC & 600    & 0.917 & 0.617 & 0.673 & 0.75 & 0.361 & 0.445 & 0.556 & 0.219 & 0.294 & 0.583 & 0.222 & 0.297\\[1pt]
CC & 750    & 0.917 & 0.577 & 0.648 & 0.75 & 0.284 & 0.371& 0.583 &0.253 &0.324 &0.472 &0.19 &0.252\\[3pt]
\cline{3-14}\\[-3pt]
%%%%%%%%%%%%%%%%%%%%%%%%%%%%%%%%%%%%%%%%%%%
MF & 450 &  0.464 & 0.302 & 0.367 & 0.462 & 0.26 & 0.316& 0.792 &0.351 &0.452 & 0.729 &0.357 &0.454\\[1pt]
MF & 600 &  0.41 & 0.222 & 0.28 & 0.487 & 0.234 & 0.302& 0.667 &0.294 &0.381 & 0.688 &0.317& 0.414\\[1pt]
MF & 750 &  0.372 & 0.162 & 0.207 & 0.333 & 0.183 & 0.224& 0.708 & 0.298 & 0.394 & 0.667 & 0.298 & 0.392\\[3pt]
\cline{3-14}\\[-3pt]
%%%%%%%%%%%%%%%%%%%%%%%%%%%%%%%%%%%%%%%%%%%%%%%
BP & 450 &   0.527 & 0.299 & 0.364 & 0.574 & 0.331 & 0.401& 0.556 & 0.203 & 0.282 & 0.569 & 0.223 & 0.305\\[1pt]
BP & 600 &  0.481 & 0.266 & 0.323 & 0.494 & 0.281 & 0.342& 0.497 & 0.185 &0.254 & 0.523 &0.193 &0.268\\[1pt]
BP & 750 &  0.443 & 0.231 & 0.297 & 0.451 & 0.24 & 0.297& 0.458 & 0.158 & 0.223 &0.451 &0.148 &0.211\\[3pt]
%%%%%%%%%%%%%%%%%%%%%%%%%%%%%%%%%%%%%%%%%%%%%%%%%%%%%%%%%%
& &  \multicolumn{3}{c}{\bf SVM AL}& \multicolumn{3}{c}{\bf RF AL}&  \multicolumn{3}{c}{\bf SVM AL}& \multicolumn{3}{c}{\bf RF AL}\\ [1pt]
CC & 450  &   0.463 & 0.173 & 0.235 & 0.25 & 0.088 & 0.12 &  0.222 & 0.066 & 0.095 & 0.222 & 0.056 & 0.087\\[1pt]
CC & 600  &   0.5 & 0.255 & 0.315 & 0.167 & 0.083 & 0.111& 0.111 & 0.035 & 0.047 &  0.056& 0.021 &0.03\\[1pt]
CC & 750  &   0.417 & 0.158 & 0.218 & 0.25 & 0.167 & 0.194& 0.111 &0.027 & 0.041 & 0.083 & 0.025 & 0.037\\[3pt]
\cline{3-14}\\[-3pt]
%%%%%%%%%%%%%%%%%%%%%%%%%%%%%%%%%%%%%%%%%%%%%%
MF & 450 &   0.201 & 0.077 & 0.103 & 0.256 & 0.095 & 0.13& 0.333 &0.109 &0.161 & 0.229& 0.065 & 0.1\\[1pt]
MF & 600 &   0.154 & 0.047 & 0.07 & 0.154 & 0.042 & 0.066& 0.312 &0.094 &0.142 & 0.146& 0.042 & 0.065\\[1pt]
MF & 750 &   0.129 & 0.035 & 0.063 & 0.128 & 0.034 & 0.053& 0.229 & 0.072 & 0.107 & 0.146 & 0.038 & 0.059\\[3pt]
\cline{3-14}\\[-3pt]
%%%%%%%%%%%%%%%%%%%%%%%%%%%%%%%%%%%%%%%%%%%%5 
BP & 450 &  0.257 & 0.129 & 0.163 & 0.278 & 0.132 & 0.17& 0.261&  0.062 & 0.096 &0.144 &0.033 &0.052\\[1pt]
BP & 600 &  0.247 & 0.111 & 0.144 & 0.228 & 0.098 & 0.129& 0.157 &0.033 &0.053 & 0.092 & 0.018 & 0.029\\[1pt]
BP & 750 &  0.227 & 0.099 & 0.121 & 0.21 & 0.084 & 0.113& 0.163 & 0.032 & 0.052 & 0.085 &0.015 &0.025\\[1pt]
\hline\hline\\[-5pt]    
\end{tabular}
\caption{Average results of SVM and RF trained using labels $\by$ for predicting proteins in $V_{-+}$. The predictions are evaluated using the newer release $\overline\by$.}\label{tab:yeast_svm_rf_cnp} 
\normalsize
\end{table}
%%%%%%%%%%%%%%%%%%%%%%%%%%%%%%%%%%%%%%%%%%%%%%%%%%%%%%%%%%%%%%%%%
%%%%%%%%%%%%%%%%%%%%%%%%%%%%%%%%%%%%%%%%%%%%%%%%%%%%%
\begin{table}[!t]
\centering
\small
\begin{tabular}{lccc}
  \hline
{\tt Method} & {\tt B} & {\tt SVM} & {\tt RF}\\ 
  \hline\\[-7pt]
PnoSub & All & 23.40 & 65.11 \\[3pt]
\cline{1-1}\\[-5pt]
PrndSub & 450 & 9.76 & 52.49\\[1pt]
PrndSub & 600 & 10.22 & 53.90 \\[1pt]
PrndSub & 750 & 11.57 & 54.74\\[3pt]
\cline{1-1}\\[-7pt]
AL & 450 & 51.54 & 239.23\\[1pt]
AL & 600 & 81.40 & 329.22\\[1pt]  
AL & 750 & 109.54 & 428.16\\[1pt]
\hline\hline\\[-5pt]    
\end{tabular}
\caption{Average running time in seconds to perform a cross validation cycle on yeast data.}\label{tab:times} 
\normalsize
\end{table}   
%%%%%%%%%%%%%%%%%%%%%%%%%%%%%%%%%%%%%%%%%%%%%%%%%%%%%

Finally, the evaluation across the temporal holdout of proteins $V_{-+}$ is shown in Table~\ref{tab:yeast_svm_rf_cnp}, where we report the performance according to annotations $\overline\by$ after predicting using $\by$ as functional labeling.
 Our approach achieves slightly better results with regard to the corresponding passive model (PnoSub) in all branches (except for SVMs, MF yeast data), and largely the top results when considering all the proteins. Indeed, the results on the the remaining proteins are the same of Table~\ref{tab:yeast_svm_rf}, since $\by$ and $\overline\by$ differs solely on the instances $V_{-+}$. Nevertheless, a clear pattern does not emerge. The F results on noisy label proteins are clearly limited by the number of positives predicted by the model, since all proteins $V_{-+}$ are positive according to the newer release $\overline\by$. F values tend thereby to decrease when the budget $B$ increases. In this context, the method with the highest recall in Table~\ref{tab:yeast_svm_rf} ---i.e., PrndSub--- wins, although it is not the method which classifies best. It is worth noting that the models were not specifically trained to predict the set $V_{-+}$; accordingly, the performance of classifiers just over the noisy label proteins relies on a side-effect of our learning algorithms.  
Instead, when evaluating over all proteins, the improvements of our strategies over {PnoSub} and {PrndSub} are confirmed, or even improved, in the holdout setting with respect to the CV setting. Note that having good performance in the holdout setting is very important, since it is a more realistic scenario for AFP.
  
Finally, to have an idea of the scalability of our methods, the average running times in seconds to perform a CV cycle on yeast data using a Linux machine with Intel Xeon(R) CPU 3.60GHz and 32 Gb RAM are reported in Table~\ref{tab:times}. As expected, the running time of active SVMs increases almost linearly with the number of selected negatives with respect to the SVM PnoSub. On the other hand, RFs scale poorly, and are much slower than SVMs while exhibiting a worse performance. This motivates our choice of employing RFs just as a mean to validate the AL negative selection.
%%%%%%%%%%%%%%%%%%%%%%%%%%%%%%%%%%%%%%%%%%%%%%%%%%%%%%%%%%%%%

%%%%%%%%%%%%%%%%%%%%%%%%%%%%%%%%%%%%%%%%%%%%%%%%%%%%%%%%%%%%%
\subsection{Predicting GO protein functions}\label{sub:afp_comp}
We compared our approach to eight state-of-the-art algorithms for AFP operating in the same flat setting (that is, not including further information from other functions in the hierarchy when a given function is predicted): \textit{Random Walk} (RW) and the \textit{Random Walk with Restart} (RWR)~\citep{Tong08}; the \textit{guilt-by-association} (GBA) method~\citep{Schwikowski00}; the \textit{Label Propagation} algorithm (\textit{LP})~\citep{Zhu03}; a method based on Hopfield networks, the \textit{Cost-Sensitive Neural Network} (COSNet)~\citep{Bertoni11}, designed for unbalanced data and showing competitive results on the MOUSEFUNC benchmark~\citep{Unipred}; the \textit{Multi-Source k-Nearest Neighbors} (MS-kNN)~\citep{Lan13}, one of the top-ranked methods in the recent CAFA2 international challenge for AFP; the \textit{RAnking of Nodes with Kernelized Score Functions} (RANKS)~\citep{Vale12c}, recently proposed as a fast and effective ranking algorithm for AFP. Finally, to assess the benefit of the negative selection, we tested also the baseline passive SVM (PnoSub).
For COSNet and RANKS we used publicly available R libraries~\citep{Frasca17,RANKS},
whereas for the other methods the code provided by the authors or our own software implementations was utilized. Since COSNet is a binary classifier, to provide a ranking of proteins we followed the approach presented in~\citep{Frasca13bis}, which uses the neuron energy at equilibrium as ranking score. For SVM we used the margin to rank instances.
Finally, the parameters required by the benchmarked methods were learned through internal tuning on a subset of training data.

We selected the GO terms with $10$ to $100$ annotated proteins in the most recent release. This in order to discard terms that are too generic (close to root nodes), and to ensure the availability of a few positives to train the models in a 3-fold CV setting.
The resulting datasets contain $162$, $227$, and $660$ terms for yeast, and $272$, $476$, and $1689$ terms for human, respectively, in the GO branches CC, MF and BP. 

%%%%%%%%%%%%%%%%%%%%%%%%%%%%%%%%%%%%%%%%%%%%%%%%%%%%%%%%%
\begin{figure*}[!t]
%\vspace{-0.3cm}
\begin{center}
\begin{tabular}{cc}
\hspace{-0.2cm} (a)  & \hspace{-0.1cm} (b)\\[0pt]
\hspace{-0.2cm} \includegraphics [width=0.5\textwidth] {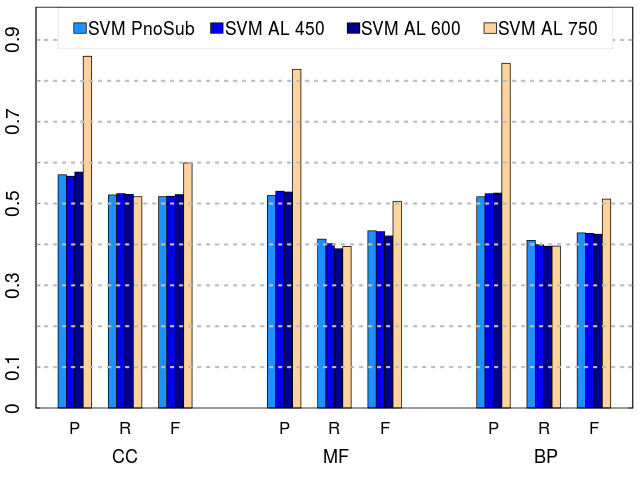}& 
\hspace{-0.1cm} \includegraphics [width=0.5\textwidth] {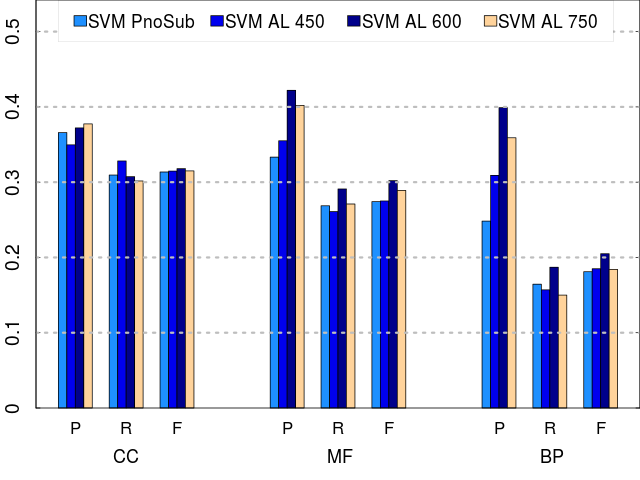}\\
\end{tabular}
\end{center}
\vspace{-0.6cm}
\caption{Precision (P), Recall(R) and F values of binary classifiers averaged across GO terms on (a) yeast and (b) human data.}\label{fig:GO_CVres}
\end{figure*}
%%%%%%%%%%%%%%%%%%%%%%%%%%%%%%%%%%%%%%%%%%%%%%%
The comparison with the passive SVM (PnoSub) is shown in Fig.~\ref{fig:GO_CVres}, where
our algorithm shows a considerable increase in both precision and $F$-measure on yeast data (Fig.~\ref{fig:GO_CVres}(a)) when the budget of negatives is large enough. Similar but less marked improvements with respect to passive SVMs are obtained on human data, except for the CC branch where the improvement in terms of F measure is not statistically significant, Fig.~\ref{fig:GO_CVres}(b). For the yeast setting, already with $B \in \{450, 600\}$ SVM AL achieves similar results compared to SVM, but using just less than $15\%$ of the available negatives. However, when $B=750$ its precision is significantly increased while nearly preserving the same recall. This results in a statistically significant improvement in the $F$-measure ($p$-$value < 0.001$), which is preferable. Indeed, in contexts characterized by a scarcity of positives, a large precision is preferable over a large recall for the same F score. In the human experiment, our method needs a lower budget ($B=600$) to achieve its top performance, which corresponds to around $5\%$ of available negatives, showing that a large proportion of negatives likely carries redundant information. When $B=750$ the results are slightly worse, likely due to overfitting phenomena. 
On both yeast and human data we tested budgets even larger than $750$, which increased the execution time without providing significant performance improvements (results not shown).   

%%%%%%%%%%%%%%%%%%%%%%%%%%%%%%%%%%%%%%%%%%%%%%%%%%%%%%%%%
\begin{figure*}[!t]
\begin{center}
\begin{tabular}{cc}
\hspace{-0.2cm} (a)  & \hspace{-0.15cm} (b)\\[0pt]
\hspace{-0.2cm} \includegraphics [width=0.5\textwidth] {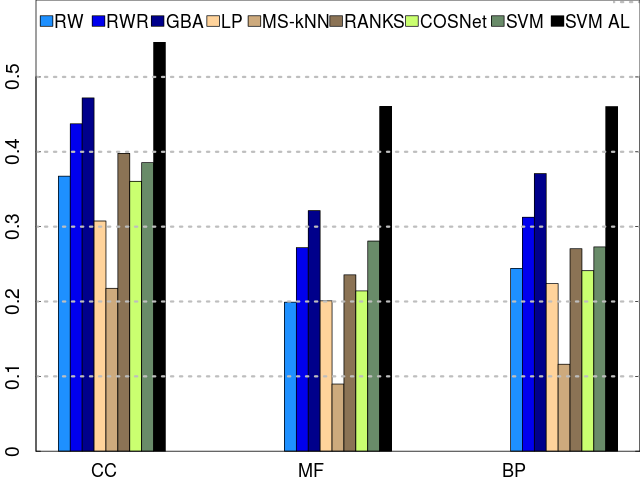}& 
\hspace{-0.15cm} \includegraphics [width=0.5\textwidth] {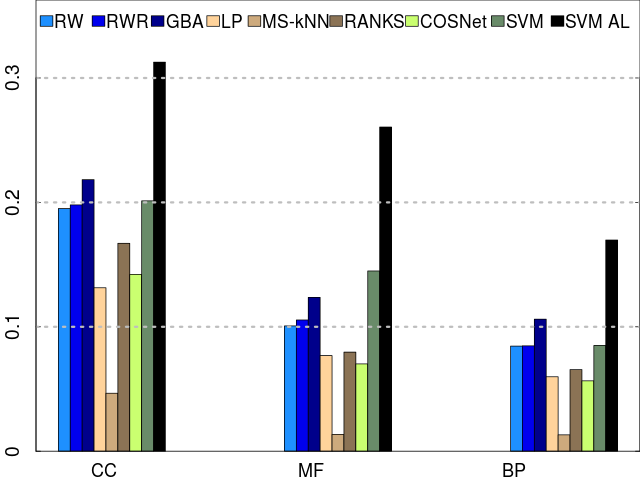}\\
\end{tabular}
\end{center}
\vspace{-0.3cm}
\caption{Performance comparison in terms of AUPR averaged  across GO terms on (a) yeast and (b) human  data.}\label{fig:GO_AUPRcomp}
\end{figure*}
%%%%%%%%%%%%%%%%%%%%%%%%%%%%%%%%%%%%%%%%%%%%%%%

\begin{figure*}[!t]
\begin{center}
\begin{tabular}{cc}
\hspace{-0.2cm} (a)  & \hspace{-0.15cm} (b)\\[0pt]
\hspace{-0.2cm} \includegraphics [width=0.50\textwidth] {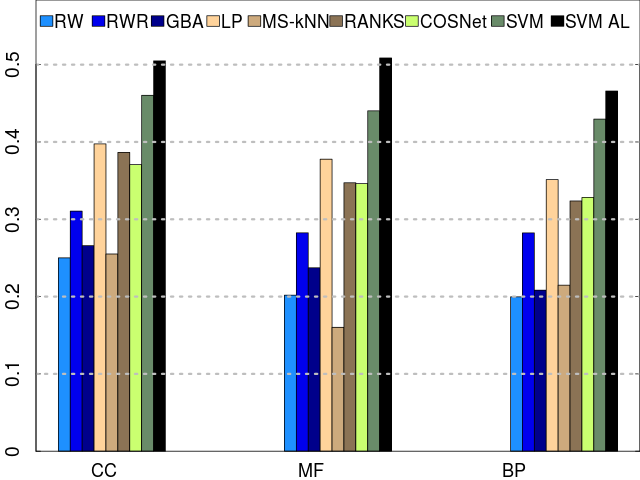}& 
\hspace{-0.15cm} \includegraphics [width=0.50\textwidth] {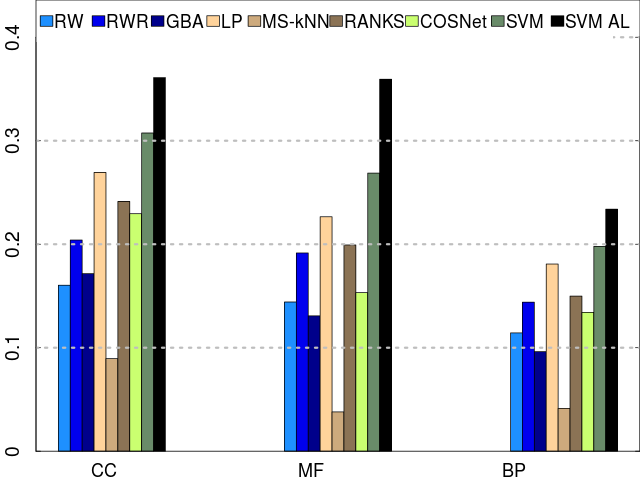}\\
\end{tabular}
\end{center}
\vspace{-0.6cm}
\caption{\textit{Fmax} results on (a) yeast and (b) human  data.}\label{fig:GO_Fmaxcomp}
\end{figure*}

The comparison with the state-of-the-art methods for AFP is depicted in Fig.~\ref{fig:GO_AUPRcomp} (AUPR) and Fig.~\ref{fig:GO_Fmaxcomp} ($\Fmax$), where $B$ is set to $750$. In terms of AUPR, SVM AL significantly outperforms all the other methods in all the branches ($p$-$value < 0.001$). SVM instead is placed between second and fifth place, disclosing the relevant contribution supplied by the negative selection procedure based on active learning. 
On the other side, the GBA algorithm behaves nicely, being the second top method in all experiments (except for human--MF), whereas RWR often is the third best performing method. The good performance of GBA and RWR confirms the results shown in~\citep{VASCON18}. MS-kNN performs worst in all the settings, followed by LP, while COSNet performs halfway between the best and worst performance.

In order to gain further insights on the behaviour of our algorithm, we followed the MOUSEFUNC challenge experimental set-up~\citep{MOUSEFUNCI} and computed the performance averaged across categories containing terms with different degrees of imbalance. Namely, we distinguished two groups: functions with $10$--$20$ positives and functions with $21$--$100$ positives. Functions in the first group are more specific (i.e., closer to leaves in the GO hierarchy), and more complex (since less information is available). Hence, on the functions in this group AFP methods typically perform worse~\citep{MOUSEFUNCI,Mostafavi10,Unipred}. Results shown in Fig.~\ref{fig:GO_AUPR_comp_lm} in the Appendix A confirm this trend, except for CC experiments where all methods show lower mean AUPR on the $21$--$100$ group ---see Fig.~(\ref{fig:GO_AUPR_comp_lm} (b)-(d)), Appendix A. SVM AL instead achieves similar results on both groups. Indeed, the gap between the two top methods is more pronounced in the $10$--$20$ group. This is likely due to the cost-sensitive strategy adopted by our method, which allows to better handle the higher imbalance of the $10$--$20$ group. Such behaviour is preferable and more useful in ontologies like GO, where more specific nodes provide less generic definitions ensuring a better characterization of protein functions.
Finally, on yeast data our algorithm tends to suffer from less outliers compared to other methods (Fig.~\ref{fig:GO_AUPR_comp_boxplots}, Appendix A), having almost the same median and mean AUPR (red horizontal segment). On human data, possibly due to the higher complexity of this dataset, such a behaviour becomes less evident.

%%%%%%%%%%%%%%%%%%%%%%%%%%%%%%%%%%%%%%%%%%%%%%%%%%%%%%%%%
%%%%%%%%%%%%%%%%%%%%%%%%%%%%%%%%%%%%%%%%%%%%%%%

SVM AL has the best performance even in terms of $\Fmax$ in all the considered settings, and especially on the MF terms for human data. SVM is the second top method.  LP performs much better with reference to AUPR results, being at the third place in all experiments and confirming the analysis proposed in~\cite{VASCON18}. COSNet and RANKS occupy mid-range positions, with a slightly better ranking on yeast data.
\section{Conclusion}
We addressed the problem of positive and unlabeled learning using a novel approach, exploiting the synergy of imbalance-aware and negative selection strategies. Our algorithm, based on Support Vector Machines, is able to counterbalance the predominance of unannotated instances via an imbalance-aware technique combined with active learning for choosing negative examples. We showed that active learning helps in both choosing reliable negatives and detecting the ``noisy'' examples (i.e., initially unannotated instances that get annotated in later releases of the dataset). The effectiveness of our approach was experimentally tested on the problem of automatically annotating the proteomes of sequenced organisms, whose proteins have annotations from the GO functional taxonomy that are sparse, and lack a proper definition for negative examples. This experimental validation showed that our tool compares favourably against existing approaches to AFP when predicting the GO functions of yeast and human proteins.

\section*{Funding}
This work was supported by the grant title \textit{Machine learning algorithms to
handle label imbalance in biomedical taxonomies}, code PSR2017$\_$DIP$\_$010$\_$MFRAS,  Universit\`a degli Studi di Milano.

%
%\begin{acknowledgements}
%\end{acknowledgements}

%\bibliographystyle{apalike}
%\bibliographystyle{plainnat}
%\bibliography{biblio,ncb}   % name your BibTeX data base

%\begin{thebibliography}{}
%%
%% and use \bibitem to create references. Consult the Instructions
%% for authors for reference list style.
%\end{thebibliography}

\appendix

\newpage
\section*{Appendix A}\label{app:A}
\ 
%\section{Predicting GO protein functions}\label{app:A}
%%%%%%%%%%%%%%%%%%%%%%%%%%%%%%%%%%%%%%%%%%%%%%%%%%%%%%%%%
\begin{figure*}[!h]
\begin{center}
\begin{tabular}{cc}
\hspace{-0.0cm} (a)  & \hspace{-0.0cm} (b)\\[0pt]
\hspace{-0.0cm} \includegraphics [width=0.5\textwidth] {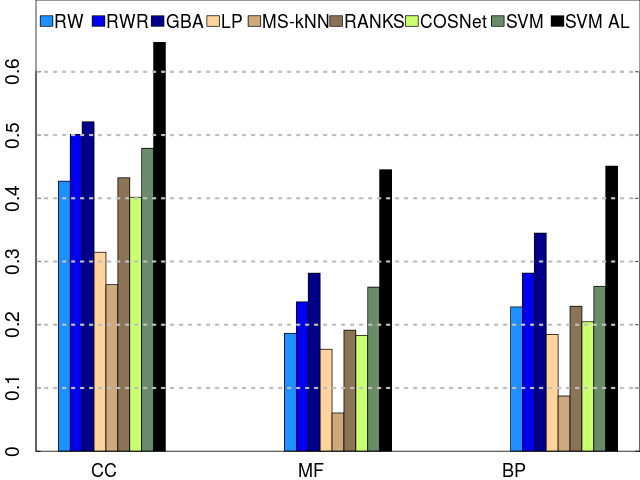}& 
\hspace{-0.0cm} \includegraphics [width=0.5\textwidth] {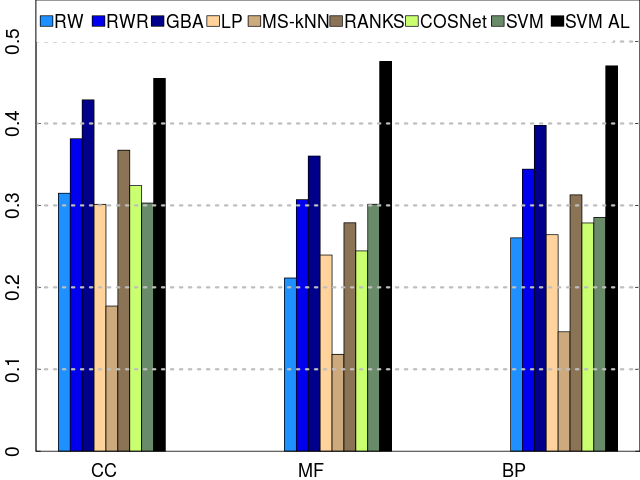}\\
\hspace{-0.0cm} (c)  & \hspace{-0.0cm} (d)\\[-0pt]
\hspace{-0.0cm} \includegraphics [width=0.5\textwidth] {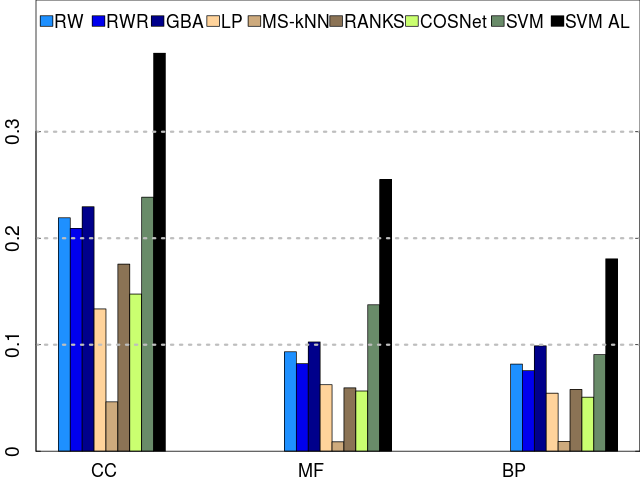}& 
\hspace{-0.0cm} \includegraphics [width=0.5\textwidth] {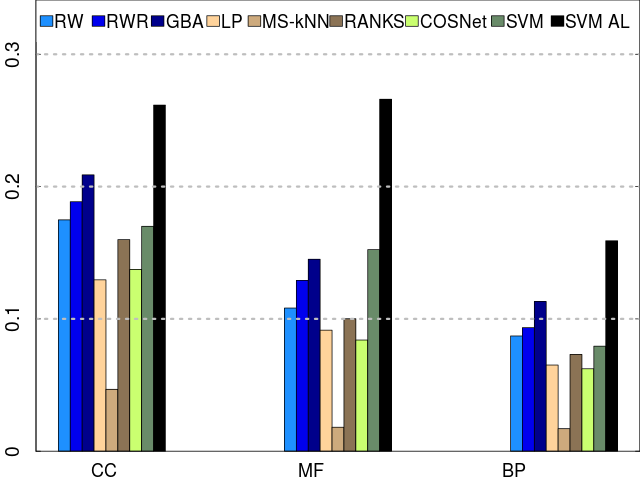}\\
\end{tabular}
\end{center}
\vspace{-0.6cm}
\caption{State-of-the-art comparison in terms of AUPR values averaged across GO terms with $10$--$20$ and $21$--$100$ positives on (a--b) yeast and (c--d) human data. First column corresponds to $10$--$20$ terms. The total number of $10$--$20$  (resp. $21$--$100$) terms is 76 (resp. 86), 122 (115) and 334 (326) for CC, MF and BP branches respectively. }\label{fig:GO_AUPR_comp_lm}
\end{figure*}
%%%%%%%%%%%%%%%%%%%%%%%%%%%%%%%%%%%%%%%%%%%%%%%

%%%%%%%%%%%%%%%%%%%%%%%%%%%%%%%%%%%%%%%%%%%%%%%%%%%%%%%%%
\begin{figure*}[!h]
\begin{center}
\begin{tabular}{cc}
\hspace{-0cm} (a)  & \hspace{-0cm} (b)\\[0pt]
\hspace{-0cm} \includegraphics [width=0.5\textwidth] {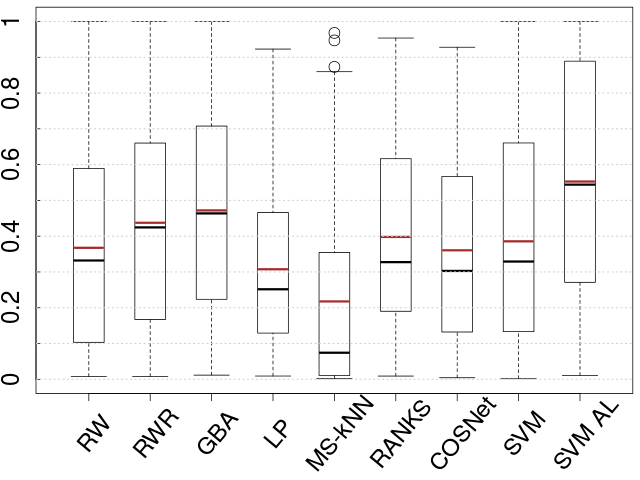}& 
\hspace{-0cm} \includegraphics [width=0.5\textwidth] {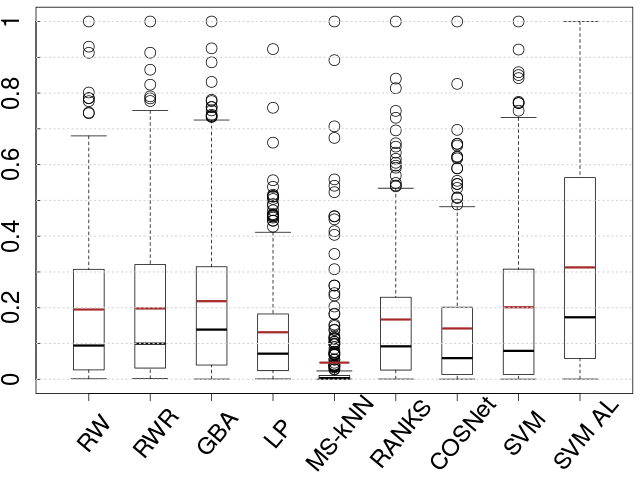}\\
\hspace{-0cm} (c)  & \hspace{-0cm} (d)\\[-0pt]
\hspace{-0cm} \includegraphics [width=0.5\textwidth] {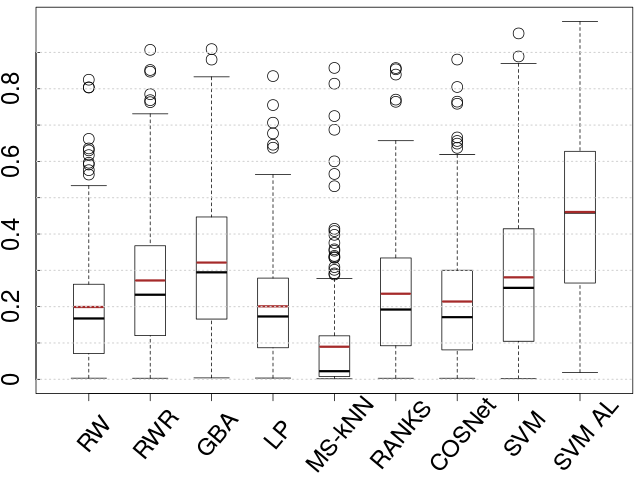}& 
\hspace{-0cm} \includegraphics [width=0.5\textwidth] {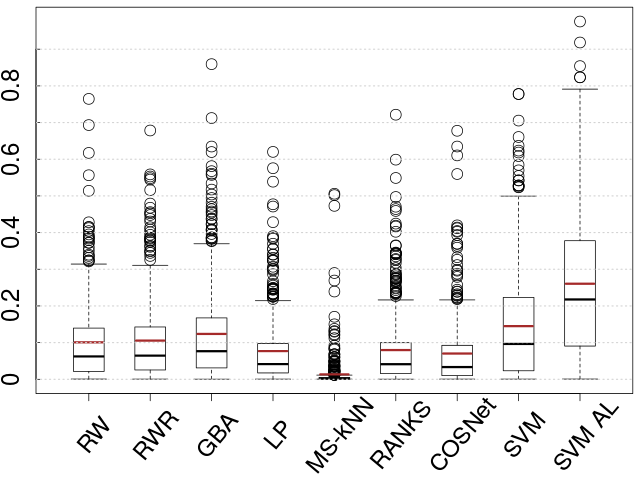}\\
\hspace{-0cm} (e)  & \hspace{-0cm} (f)\\[-0pt]
\hspace{-0cm} \includegraphics [width=0.5\textwidth] {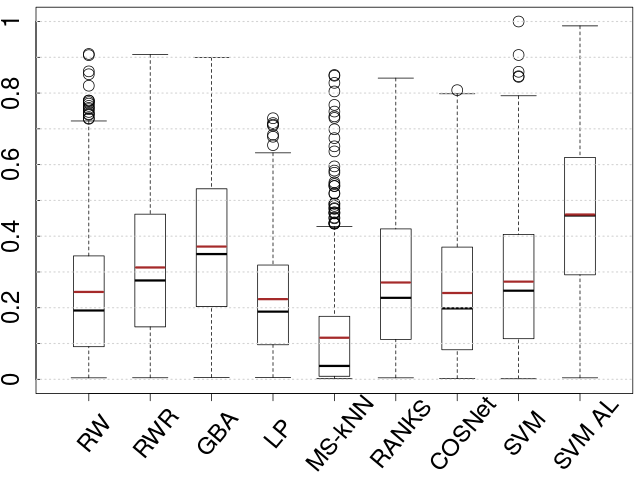}& 
\hspace{-0cm} \includegraphics [width=0.5\textwidth] {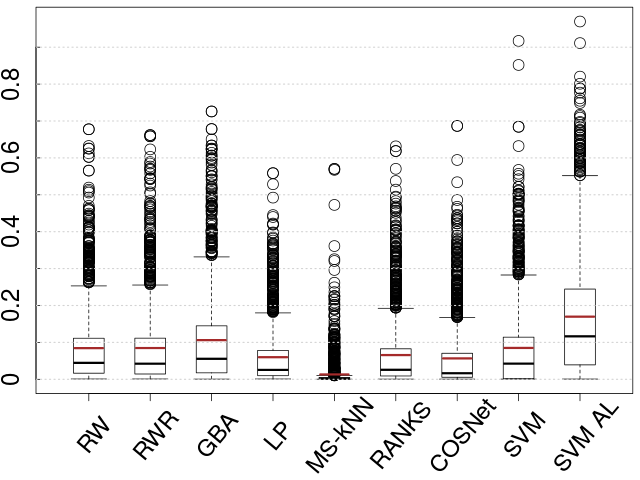}\\
\end{tabular}
\end{center}
\vspace{-0.6cm}
\caption{AUPR boxplots of the state-of-the-art comparison  on yeast (first column) and  human (second column) data. First row corresponds to CC terms, second and third rows to MF and BP terms.}\label{fig:GO_AUPR_comp_boxplots}
\end{figure*}
%%%%%%%%%%%%%%%%%%%%%%%%%%%%%%%%%%%%%%%%%%%%%%%
\end{document}